\documentclass{article}
\usepackage[preprint,nonatbib]{nips_2018}
\usepackage[utf8]{inputenc} 
\usepackage[T1]{fontenc}    
\usepackage{url}            
\usepackage{booktabs}       
\usepackage{amsfonts}       
\usepackage{nicefrac}       
\usepackage{microtype}      
\usepackage{graphicx}
\usepackage{amsmath}
\usepackage{enumerate}
\usepackage[dvipsnames]{xcolor}
\usepackage{tikz,pgf,pgfplots}
\usetikzlibrary{shapes.geometric, arrows}
\usepackage[format=plain,labelfont={bf,it},textfont=it]{caption}
\usepackage{floatrow}

\usepackage[colorlinks=true,citecolor=black!50!green,linkcolor=black!33!red]{hyperref}
\usepackage{csquotes}
\usepackage[nameinlink,noabbrev]{cleveref}

\newfloatcommand{capbtabbox}{table}[][\FBwidth]

\usepackage[backend=bibtex,style=numeric-comp]{biblatex}
\bibliography{lit}

\newcommand{\argmax}[1]{\underset{#1}{\operatorname{arg}\,\operatorname{max}}\;}

\hypersetup{
    colorlinks=true,
    linkcolor=black!33!red,,
    urlcolor=black,
    linktoc=all
}

\tikzstyle{rectangle} = [rectangle, minimum width=1cm, minimum height=1cm, text centered, text width = 2cm, draw=black, fill=orange!30]
\tikzstyle{arrow} = [thick,->,>=stealth]

\title{Application of Decision Rules for Handling Class Imbalance in Semantic Segmentation}

\author{
	Robin Chan\qquad Matthias Rottmann\qquad Hanno Gottschalk \\
  	School of Mathematics and Natural Sciences \\
  	University of Wuppertal\\
  	\texttt{\{\href{mailto:rchan@uni-wuppertal.de}{rchan},\href{mailto:rottmann@uni-wuppertal.de}{rottmann},\href{mailto:hgottsch@uni-wuppertal.de}{hgottsch}\}@uni-wuppertal.de}\\  	
  	\AND
  	Peter Schlicht\qquad Fabian Hüger \\
  	Architecture and AI Technologies \\
  	Volkswagen Group Research Automated Driving \\  
  	\texttt{\{\href{mailto:peter.schlicht@volkswagen.de}{peter.schlicht},\href{mailto:fabian.hueger@volkswagen.de}{fabian.hueger}\}@volkswagen.de} \\
}

\begin{document}

\maketitle

\begin{abstract}
As part of autonomous car driving systems, semantic segmentation is an essential component to obtain a full understanding of the car's environment. One difficulty, that occurs while training neural networks for this purpose, is class imbalance of training data. Consequently, a neural network trained on unbalanced data in combination with maximum a-posteriori classification may easily ignore classes that are rare in terms of their frequency in the dataset. However, these classes are often of highest interest. We approach such potential misclassifications by weighting the posterior class probabilities with the prior class probabilities which in our case are the inverse frequencies of the corresponding classes in the training dataset. More precisely, we adopt a localized method by computing the priors pixel-wise such that the impact can be analyzed at pixel level as well. In our experiments, we train one network from scratch using a proprietary dataset containing 20,000 annotated frames of video sequences recorded from street scenes. The evaluation on our test set shows an \emph{increase of average recall} with regard to instances of pedestrians and info signs by $25\%$ and $23.4\%$, respectively. In addition, we significantly \emph{reduce the non-detection} rate for instances of the same classes by $61\%$ and $38\%$.
\end{abstract}

\section{Introduction} \label{seg:introduction}

A common issue with ``real world'' datasets is the imbalance of observed object classes. Class imbalance in datasets can have a detrimental effect on classification performance of neural networks (NNs) trained on such datsets, see also \cite{Buda17}. Methods overcoming class imbalance can be divided into two main categories \cite{Small17}. The first category are \emph{sampling-based} methods that operate directly on a dataset with the aim to balance its class distribution. Oversampling and undersampling strategies have been proposed in \cite{Small17,Buda17,More16}. In their basic versions, the dataset is balanced by increasing the number of instances from ``minority'' classes and by decreasing the number of instances from ``majority'' classes, respectively. A more advanced method called SMOTE \cite{Bowyer11} combines these two aforementioned approaches, resulting in an additional boost in classification performance.

The second category are \emph{algorithm-based} methods \cite{Small17,Buda17,Lopez13}. They make use of cost-based training and decision thresholding. The idea behind these strategies is to assign different costs to classification mistakes for different classes. Accordingly, one possibility is to minimize the misclassification cost instead of the standard loss function \cite{Kukar98} during training. This would however bias the softmax probability output of the NN. The other possibility is to make class predictions cost-sensitive during inference phase after the network is fully trained by moving the output threshold towards inexpensive classes, see \cite{zhou06}.

\begin{figure}[]
\centering
\vspace{-1cm}
\input{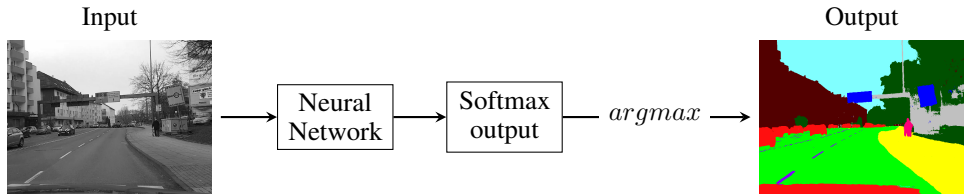}
\caption{Illustration of semantic segmentation.}
\label{fig:sem-seg}
\end{figure}

In this work we deal with a proprietary semantic segmentation dataset of the Volkswagen Group and convolutional neural networks (CNNs), more precisely a Full Resolution Residual Network (FRRN) \cite{Pohlen16}, trained on this dataset. In contrast to the publicly available Cityscapes dataset \cite{cityscapes16}, our dataset also contains scenes from highways and country roads. Consequently, classes like humans and traffic signs are underrepresented. Using a CNN trained for this task, the semantic segmentation as a pixel-wise classification is obtained by the maximum a-posteriori probability (MAP), i.e., by applying the $\mathop{argmax}$ function to the pixel-wise softmax output, see \cref{fig:sem-seg}.
The CNN as a statistical model aims at minimizing the chance of a misclassification which in decision theory is known as Bayes rule. Another mathematically natural approach from decision theory is the \textit{Maximum Likelihood} (ML) rule. While the MAP / Bayes rule incorporates a prior belief about the semantic classes, the maximum likelihood rule decides only by means of the observed features and chooses the most typical class for the given pattern. From now on, we use the abbreviations \emph{Bayes} and \emph{ML} to refer to these decision rules.

Our approach of using the ML decision rule is motivated by work that studies the influence of risk factors on heart diseases \cite{Fahrmeir-1996,kleinbaum78}. Given a person's features a decision function is computed to determine whether a patient suffers from a heart disease or not. While the total number of falsely diagnosed patients is increased when using ML instead of Bayes, the number of falsely diagnosed patients, who are actually ill, is significantly reduced. This follows from the substantially rare occurrence of patients with a heart disease that the Bayes rule assumes as a prior belief. The ML rule determines the disease independently of this assumption.

Class balancing is not widely applied to CNNs for semantic segmentation. For instance, class balancing is taken into consideration in Fully Convolutional Networks (FCNs) \cite{Long16}, however the authors decide not to take action since the training data is only moderately unbalanced. Furthermore, in SegNet \cite{Badrinarayanan15} \emph{median frequency balancing} is used, i.e., the weight assigned each class is the corresponding inverse class frequency multiplied by the median of all class frequencies over the whole dataset. It has been shown empirically that using the computed weights in the loss function during training results in a sharp increase in class average accuracy and a slight increase in mean intersection over union, whereas the global accuracy decreases \cite{Badrinarayanan15}.

With our dataset we pursue the approach to cover a wide variety of everyday street scenes in a preferably unbiased fashion. Thus, we do not want to change the training procedure, i.e., we neither change the loss function used for training nor balance the dataset with respect to the classes. However, we cannot ignore the issue of class imbalance and propose to approach this problem by applying decision rules.

In this work, we analyze the impact of applying the ML instead of the Bayes decision rule for CNNs for semantic segmentation. In other words, the dataset and the training procedure remain unchanged and the decision rules are only interchanged at inference. In contrast to \cite{rottmann18} that deals with false positive predictions, our main focus is to reduce false negative predictions. The remainder of this work is structured as follows: In \cref{sec:decision-rules-statistics} we explain decision rules in general and in \cref{sec:decision-rules-nn} we employ them in combination with neural networks. Next, we implement ML and evaluate its performance in particular in comparison with Bayes in \cref{sec:experiments}. For our experiments, we train one FRRN from scratch on our dataset called ``DS20k'' containing $20,000$ annotated images of traffic scenes in Europe. The DS20k dataset is highly unbalanced, especially with respect to the classes ``person'' and ``info sign'' that are significantly underrepresented compared to the remaining classes. This setting is very different from the setting in the \emph{Cityscapes dataset} \cite{cityscapes16} where these two classes are naturally boosted due to all recorded images showing urban street scenes. Moreover, we apply ML at pixel level in order to handle class imbalance in a position-specific manner.

\section{Decision rules in discriminant analysis} \label{sec:decision-rules-statistics}

Discriminant analysis is a multivariate statistical analysis task. Given a population consisting of two or more pre-defined clusters, in which each element of the population belongs to exactly one cluster, one wishes to classify observed data into these distinct groups. Therefore, the objective of discriminant analysis is to find a function that discriminates objects of the population based on observable features, and to predict the object's class affiliation from it.

Let $\Omega$ be a population consisting of $N \geq 2$ disjoint subsets. For each element $\omega \in \Omega$ we assume there exists one feature vector $x(\omega) \in S \subset \mathbb{R}^n$. Let
\begin{align*}
    \phantom{\hat{k}}
    X&:\Omega \to S\ \\ 
    K&:\Omega \to \{1,\ldots,N\} \phantom{\hat{k}}
\end{align*}
be random variables for feature vector $x$ and class affiliation $k$, respectively.
Then, we define a \textit{decision rule}
\begin{align*}
\begin{split}
d : \ S &\to \{ 1, \ldots ,N \} \\ 
x(\omega) &\mapsto \hat{k}
\end{split}
\end{align*}
to be a function which assigns an element from the feature space to one class. We say, $d(x)=\hat{k}$ is the predicted class for feature vector $x$. Furthermore, we describe
\begin{enumerate}[(i)]
\item the a-priori probability of a random object to belong to class $k$ as
\begin{equation} \label{eq:aprio}
p(k) := P(K=k) > 0,
\end{equation}
\item the a-posteriori probability of an object to belong to class $k$ given feature $x$ as
\begin{equation} \label{eq:apost}
p(k|x) := P(K=k|X=x) = \frac{P(K=k,X=x)}{P(X=x)}
\end{equation}
\item and the conditional likelihood of an object of class $k$ having feature $x$ as  
\begin{equation} \label{eq:conp}
p(x|k) := P(X=x|K=k) = \frac{P(K=k,X=x)}{P(K=k)} = \frac{P(K=k|X=x)P(X=x)}{P(K=x)}.
\end{equation}
\end{enumerate}

Usually, none of these probabilities are known and they all need to be estimated. Assuming that this is already accomplished, we define the following decision rules:

The \textit{Bayes} decision rule maps feature vectors $x$ to the class that gives the largest a-posteriori probability. Thus, the decision rule is defined by:
\begin{equation}\label{eq:bayesrule}
d_{\textit{Bayes}}(x)=\argmax{k\in\{1,\ldots,N\}}p(k|x).
\end{equation}
On the contrary, the \textit{Maximum Likelihood} decision rule maps feature vectors $x$ to the class with the largest conditional likelihood. Thus, the decision rule is defined by:
\begin{equation} \label{eq: ml-rule}
d_{\textit{ML}}(x)=\argmax{k\in\{1,\ldots,N\}}p(x|k) \stackrel{(\ref{eq:conp})}= \argmax{k\in\{1,\ldots,N\}}p(k|x)p(x)/p(k)= \argmax{k\in\{1,\ldots,N\}}p(k|x)/p(k).
\end{equation}
In the latter, the class affiliation $k$ is an unknown parameter that needs to be estimated using the principle of maximum likelihood. The decision rule $d_{\textit{ML}}(x)$ aims at finding the class $k$ for which the features $x$ are most typical (according to the observed features in the training set), independent of the a-priori probability of the particular class. The difference between these two decision rules lies in the adjustment with the prior class probabilities $p(k)$. Obviously, both decision rules are equal when the prior class distribution is balanced, i.e., $p(1)=\ldots=p(N)$.

\section{Decision rules in neural networks for semantic segmentation} \label{sec:decision-rules-nn}
Let $x \in \{0,\ldots,255\}^{m \times n}$ be a (gray-scaled) input image with resolution $m \times n$. In analogy with the previous section, the pixel-wise classification in semantic segmentation is then performed by the $\mathop{argmax}$ function for the estimated class probabilities $p_{ij}(k|x)$ for classes $k\in\{1,\ldots\, N \}$, where $(i,j) \in \{1,\ldots,m\} \times \{1,\ldots,n\}$ corresponds to the pixel position in the input image and the values for $p_{ij}(k|x)$ are obtained from the softmax output of a segmentation network. This procedure maximizes the overall probability for a correct class estimation which is equivalent to the Bayes rule in decision theory. As stated in \cite{Fahrmeir-1996}, this decision rule is optimal for the symmetric cost function 
\begin{equation}
c_s\left(\hat{k},k\right) := 
\begin{cases}
0 &,\ \text{if}\quad \hat{k}=k \\
C &,\ \text{if}\quad \hat{k} \neq k
\end{cases}\ ,\
C \in \mathbb{R}^+:=(0,\infty)
\end{equation}
with $\hat{k}$ being the predicted class while $k$ being the target class. This function implies an equal class weighting, also weighting every confusion of two classes, i.e., each type of misclassification, equally. In contrast to that, the ML rule is optimal for the inverse proportional cost function
\begin{equation}
c_p\left(\hat{k},k\right) := 
\begin{cases}
0 &,\ \text{if}\quad \hat{k}=k \\
C/p(\hat{k}) &,\ \text{if}\quad \hat{k}\neq k
\end{cases}\ ,\
C \in \mathbb{R}^+
\end{equation}
which increases the cost of a misclassification if the a-priori class probability (in the following called \emph{prior}) is low.
Consequently, we need to first determine the class distribution of our dataset in order to apply ML instead of Bayes.

\subsection{Class imbalance of data}
\begin{figure}
\begin{floatrow}
\ffigbox{%
  \centerline{\includegraphics[trim={0 0.5cm 0 1.5cm},width=\linewidth]{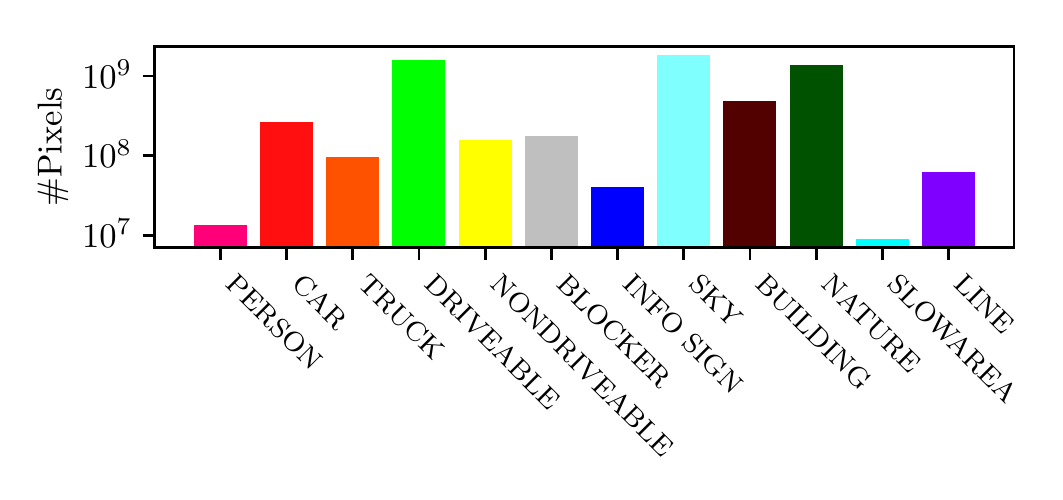}}
}{%
  \caption{Illustration of the class imbalance of the DS20k training dataset, for each class we state the number of affiliated pixels in the whole dataset.}%
  \label{fig:ds20k-imbalance}
}
\ffigbox{%
   \centerline{\includegraphics[trim={0 0.5cm 0 1.5cm},width=\linewidth]{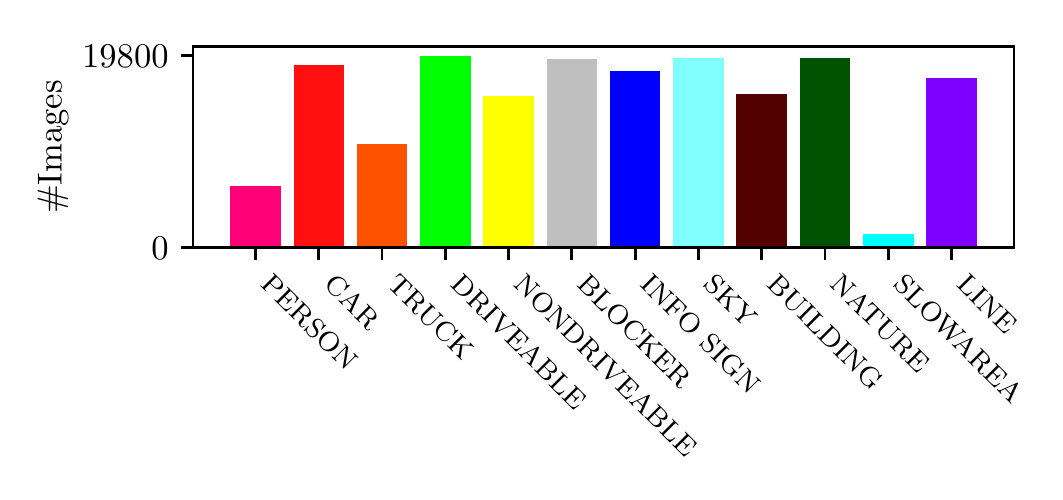}}
}{%
  \caption{Illustration of the class imbalance of the DS20k training dataset, for each class we state the number of images in the dataset containing at least one object of the given class.}%
  \label{fig:ds20k-imbalance-app}
}
\end{floatrow}
\end{figure}

A statistical analysis of our dataset reveals an unbalanced class distribution in the training set that differs significantly from a uniform distribution, cf.\ \cref{fig:ds20k-imbalance} and \cref{fig:ds20k-imbalance-app}.
For instance, the total number of pixels in \textit{DS20K} belonging to the class PERSON is $1.3 \cdot 10^7$, whereas $1.5 \cdot 10^9$ pixels belong to the class DRIVEABLE. That is a difference in two orders of magnitude. The confusion of these two classes would lead to possibly fatal situations and should be avoided, especially in the domain of near field perception. For the evaluation of the different decision rules, we compute priors and apply the decision rules at pixel level. Note, that the used segmentation network in our experiments will be a fully convolutional neural network which, on an input of infinite size, conserves the translation invariance of convolutional layers. For fully convolutional networks with a receptive field which is small compared to the dimensions of the image, an averaging of pixel-wise priors along the orbits of the translation group (adequately coarsened by pooling) would be adequate \cite{Cohen16}. However, the receptive field of our network for a single output pixel contains up to $2/3$ of the input image. Thus, almost all output pixels are affected by boundary effects which enables them to guess their approximate location and justifies our decision for pixel-wise priors. For further discussion we refer to the appendix.

\subsection{Computing priors}

\begin{figure}
\vspace{-.5cm}
\begin{floatrow}
\ffigbox{%
  \centerline{\includegraphics[trim={0 0.5cm 0 0.7cm},width=.49\textwidth]{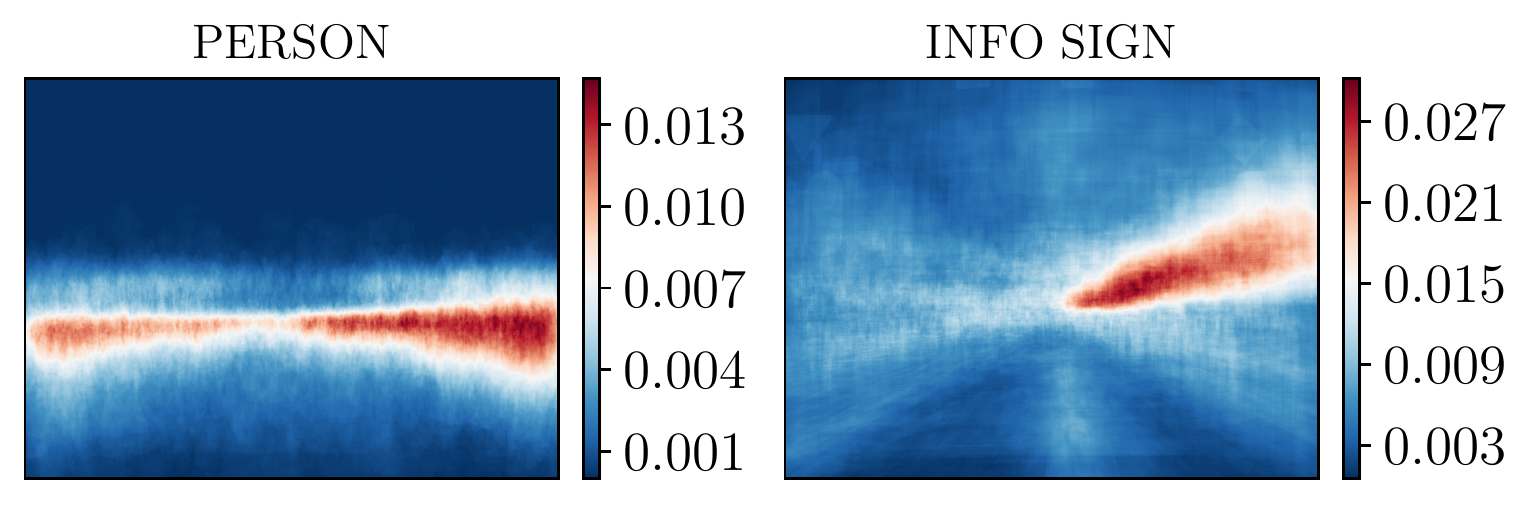}}
}{%
  \caption{Pixel-wise priors for class PERSON and INFO SIGN. For determining the priors, the frequency of class appearance at every pixel position in the training set is divided by the total number of training images.}%
  \label{fig:priors-heat}
}
\ffigbox{%
  \centerline{\includegraphics[trim={0 0.5cm 0 0.7cm},width=.49\textwidth]{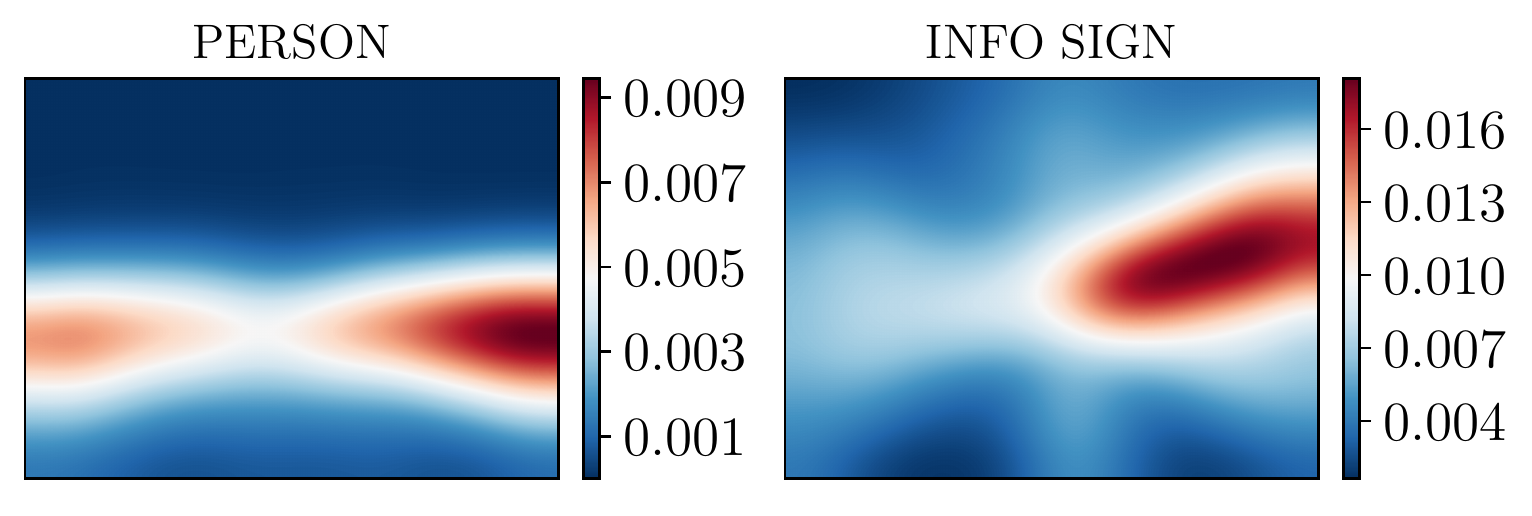}}
}{%
  \caption{Smoothed pixel-wise priors for class PERSON and INFO SIGN. For smoothing the priors, a Gaussian-filter is applied with parameter $\sigma=80$ on the actual priors in order to reduce training data specific noise.}%
  \label{fig:priors-heat-smooth}
}
\end{floatrow}
\end{figure}

The priors are essential for the implementation of decision rules. We approximate them using the training set since our network is trained on these unbalanced data. \Cref{fig:ds20k-imbalance} shows the class distribution on full image level in the entire dataset. As there are image regions, where it is more likely that a certain class appears, we are interested in the pixel-wise class distribution of the training dataset. From \cref{fig:priors-heat} we conclude, that during training there are no pedestrians seen in the upper part of the image. The network thus will be biased towards not predicting a person in that area which might be wrong, e.g., when the street is ascending.

In order to reduce training data specific noise and details from the priors, we smoothen them using a Gaussian filter. Also, a lower cut off limit of $10^{-5}$ is applied to the priors of all classes, for the purpose of avoiding divisions by zero, see \cref{eq: ml-rule}. A visualization can be found in \cref{fig:priors-heat-smooth}.

\section{Numerical experiments with decision rules} \label{sec:experiments}
We evaluate the performance of Bayes and ML on $200$ annotated test images from DS20k that were not used during training. The network we use in our experiments is a slightly modified version of the full-resolution residual network (FRRN) \cite{Pohlen16} which is a combination of SegNet \cite{Badrinarayanan15} and ResNet \cite{HeZRS15} and performs well with respect to recognition and localization.
Unlike the original version of the network we use dropout regularization \cite{JMLR14} instead of batch normalization. Furthermore, we modify the number of channels per layer.

We implement our FRRN with Tensorflow \cite{tensorflow2015-whitepaper} and train the network by minimizing the negative log-likelihood loss function using the ADAM optimizer \cite{Ruder16}. Besides that, we train the network on $19,\!800$ annotated training images (resolution $640\times480$ pixels) of traffic scenes with $12$ different semantic classes. We use a batch size of $16$, resulting in a training time of $1$ day and $16$ hours for $100$ epochs with $4$ NVIDIA Geforce GTX $1080$Tis.

\subsection{Visual comparison of Bayes and Maximum Likelihood decision rule}
\begin{figure}
\begin{floatrow}
\ffigbox{%
  \begin{minipage}{.25\textwidth}
  \centering
  \includegraphics[width=.9\textwidth]{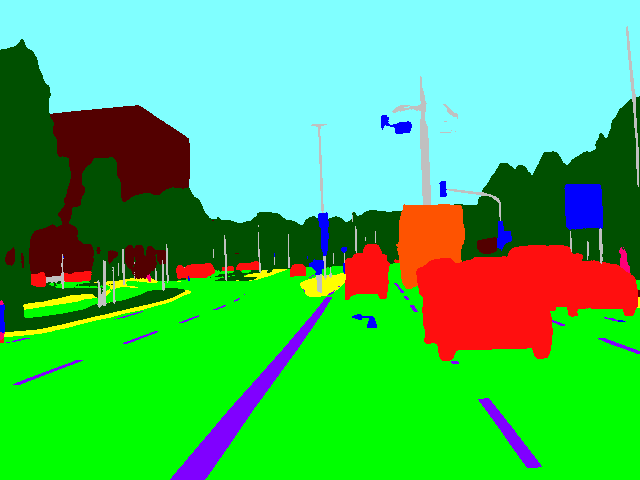}
\end{minipage}%
\begin{minipage}{.25\textwidth}
  \centering
  \includegraphics[width=.9\textwidth]{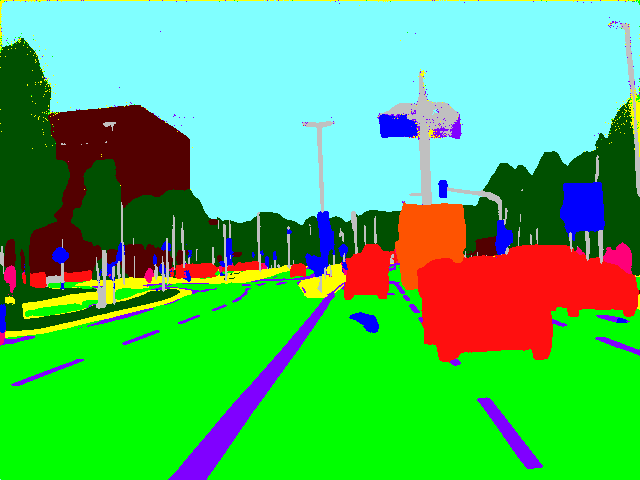}
\end{minipage}
}{%
\caption{Visualization of a segmentation with Bayes (Left) and ML (Right). The Bayes prediction serves as benchmark. The ML prediction provides additional insights about the network's segmentation capabilities. Especially the points, where different classes are predicted and thus the decision rules disagree, will be of interest for further analysis.}%
\label{fig:seg-comp}
}
\ffigbox{%
\begin{minipage}{.25\textwidth}
  \centering
  \includegraphics[width=.9\textwidth]{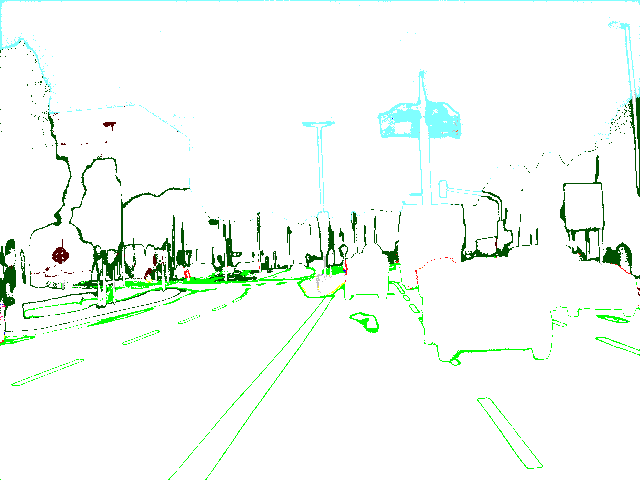}
\end{minipage}%
\begin{minipage}{.25\textwidth}
  \centering
  \includegraphics[width=.9\textwidth]{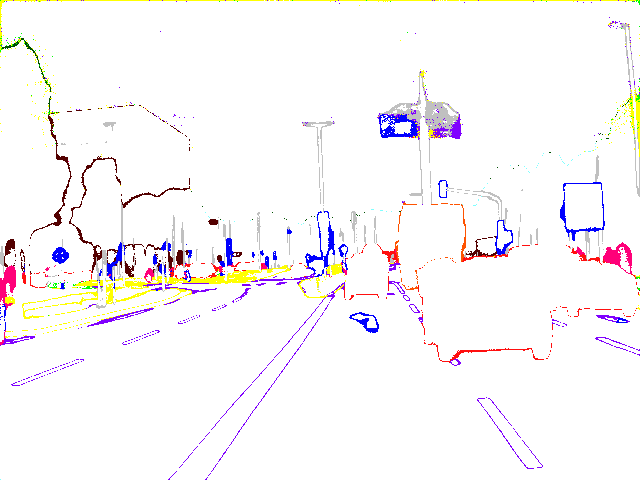}
\end{minipage}
}{%
\caption{Visualization of the differences between the segmentations obtained from Bayes and ML. In both images the predictions differ mostly at the object boundaries, i.e., at the transition from one class to another. Left: The points in the Bayes prediction where Bayes and ML disagree. Right: The points in the ML prediction where Bayes and ML disagree.}%
\label{fig:seg-dif}
}
\end{floatrow}
\end{figure}

Before we start a quantitative analysis on the impact of ML we visualize the segmentations obtained by both decision rules in order to get a basic understanding of the differences, see \cref{fig:seg-comp}.

At first glance we observe that there are no significant differences between both segmentations. Both decision rules produce the same output for most of the pixels which indicates that the network is well-trained and predicts with high confidence. In particular, the predictions for frequent classes, such as road, nature, sky and buildings, enhance this impression because they make up a large portion of the image. Therefore, the pixel positions, where the decision rules produce different predictions, are of special interest for us.

On closer inspection we observe that the Bayes and ML predictions differ remarkably often at the transition from one class to another. While Bayes prefers to predict the more frequent class at these pixel positions, ML does the opposite and prefers the less frequent class. Thus, ML enlarges the size of ``minority'' class objects compared to Bayes for the purpose of decreasing the risk of missing any pixels belonging to rare classes. Moreover, we observe that ML produces many (in terms of class affiliation) isolated false positive pixels which are boosted by the priors. For instance, in \cref{fig:seg-dif} ML frequently classifies scattered pixels as NONDRIVEABLE (pavement, traffic island,...) in the upper part of the image. Since the class $k=\mathrm{NONDRIVEABLE}$ has an extremely small a-priori probability in the upper part of the image, we see that even small posterior probabilities $p(k|x)$ can result in $p(k|x)/p(k)$ being dominant and thus $k$ being the predicted class when using the ML decision rule. In order to reduce uncertainty, we employ Monte-Carlo dropout, see \cite{Gal:2016:DBA:3045390.3045502,DBLP:journals/corr/KendallBC15}, at inference by computing $10$ different predictions under dropout and averaging the output probabilities.

Due to the occurrence of very local misclassifications we add two post processing steps: First, we discard all connected components of one class that contain less than $10$ pixels in our statistical computations. Second, we treat connected components of the same class which have less than $10$ pixels in-between as one connected component.

\subsection{Experiments with Bayes and Maximum Likelihood decision rule}
Let $A_{k,\hat{k}} \in \mathbb{N}_0$ be the number of pixels of class $k$ predicted to belong to class $\hat{k}$. For the evaluation, we compute three different performance measures that are common in semantic segmentation:

\begin{enumerate}[(i)]
\item Precision
\begin{equation} \label{eq:precision}
prc_j = A_{j,j} / \textstyle\sum_{k=1}^N A_{k,j},\ j=1,\ldots,N
\end{equation} 
\item Recall
\begin{equation} \label{eq:recall}
rec_j = A_{j,j} / \textstyle\sum_{k=1}^N A_{j,k} ,\ j=1,\ldots,N
\end{equation}
\item Intersection over Union
\begin{equation} \label{eq:iou}
IoU_j = A_{j,j}/ \left( A_{j,j} + \textstyle\sum_{k=1,k \neq j}^N (A_{k,j} + A_{j,k}) \right) ,\ j=1,\ldots,N.
\end{equation}
\end{enumerate}

\paragraph{Mean Intersection over Union.}
We compare the segmentation performance of Bayes and ML, first in terms of \textit{mean intersection over union} (mIoU) which is the average value over all classes for the intersection over union (IoU), i.e., $mIoU = \sum_{k=1}^N IoU_{k} / N$.

\begin{figure}
\vspace{-.5cm}
\begin{floatrow}
\ffigbox{%
  \centerline{\includegraphics[trim={0 0.5cm 0 0.5cm},width=.49\textwidth]{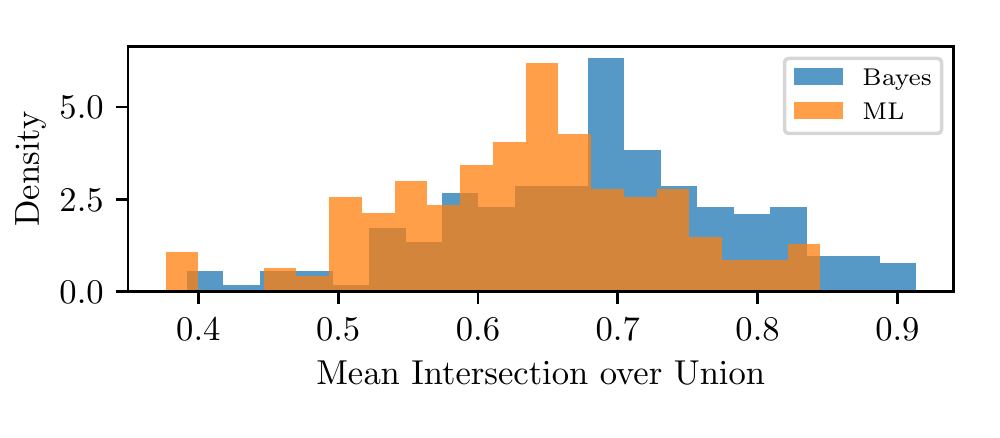}}%
}{%
  \caption{Histogram of mIoU scores considering all 200 test images}%
  \label{fig:hist-miou}
}
\capbtabbox{%
\scalebox{0.93}{
\begin{tabular}{l|lll}
\toprule
Decision rule & Overall & PERSON & INFO \\ \midrule
Bayes & 68.8 & 40.7   & 38.7 \\
ML    & 63.6 & 22.3   & 29.5 \\ \bottomrule
\end{tabular}
}
\vspace{.7cm}
}{\caption{mIoU scores (in percent) additionally averaged over all images.}%
\label{table:miou}
}
\end{floatrow}
\end{figure}

\Cref{fig:hist-miou} shows that Bayes is superior to ML regarding the overall performance. In nearly all images the mIoU for Bayes is higher than for ML. Further averaging of the mIoU over all test images in \cref{table:miou} reveals a difference of more than $5\%$ in mIoU. This finding is not unexpected since Bayes maximizes the overall probability of correct class predictions. We also compare the IoUs for the rare classes PERSON and INFO SIGN (short hand: INFO). They even show an increased superiority of Bayes. In the subsequent paragraphs we show that this is indeed caused by an overproduction of false positives when using ML.

\paragraph{Precision vs. Recall.}

\begin{figure}
\begin{floatrow}
\capbtabbox{%
\scalebox{0.93}{
\begin{tabular}{l|lll}
\toprule
Class	& Bayes	& ML & GT \\ \midrule
PERSON 	& 289 	& 857	& 368 \\
INFO	& 686	& 1592	& 1349 \\ \bottomrule
\end{tabular}
}
}{%
  \caption{Number of connected components of class PERSON and INFO.}
  \label{table:cc}%
}
\capbtabbox{%
\scalebox{0.93}{\begin{tabular}{l|cc|cc|cc}
\toprule
       & \multicolumn{2}{|c}{Precision} & \multicolumn{2}{|c}{Recall} & \multicolumn{2}{|c}{IoU} \\ \midrule
Class  & Bayes          & ML           & Bayes        & ML          & Bayes       & ML        \\ \midrule
PERSON & 61.1           & 37.4         & 48.1         & 73.1        & 31.7        & 27.5      \\ 
INFO   & 67.6           & 36.9         & 33.6         & 57.0        & 23.4        & 21.2      \\ \bottomrule
\end{tabular}}
}{\caption[Average evaluation score of a connected component]{Average precision, recall and IoU scores of a connected component in prediction}%
}
\end{floatrow}
\end{figure}

Since ML produces many additional false positives for rare classes, we would hope in this case to obtain less false negatives with ML compared to Bayes.
By comparing the total number of connected components in the differently predicted segmentations and in the corresponding ground truth segmentation, see \cref{table:cc}, we immediately take notice of a significant impact of ML: the number of connected components in ML segmentations exceeds the number of connected components in ground truth (GT) segmentations for both, the PERSON and INFO classes. In contrast to this, Bayes overlooks a significant amount of components.

Consequently, as we expect whole instances to be false positive in ML segmentations but also to recognize more of the rare objects compared to Bayes, we now analyze segment-wise precision and recall in more detail.
For this purpose, we define that a (selection of) connected component(s) predicted by some decision rule is a correct object prediction, if there is a ground truth connected component of the same class with non-empty intersection.

\begin{figure}
\vspace{-.7cm}
\begin{floatrow}
\ffigbox{%
  \centerline{\includegraphics[trim={0 0.5cm 0 0},width=.49\textwidth]{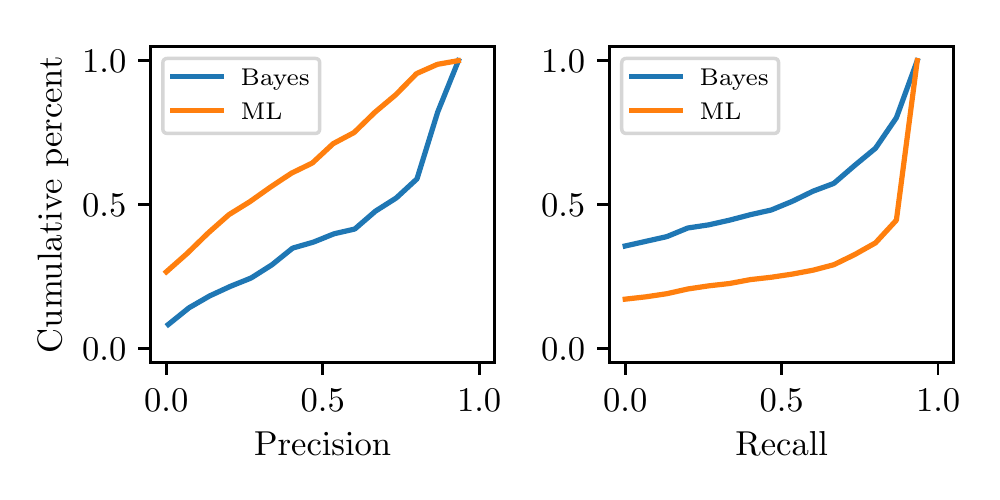}}
}{%
  \caption{Cumulative distribution function for precision and recall of class PERSON}%
  \label{fig:cdf-pers}
}
\ffigbox{%
  \centerline{\includegraphics[trim={0 0.5cm 0 0.5cm},width=.49\textwidth]{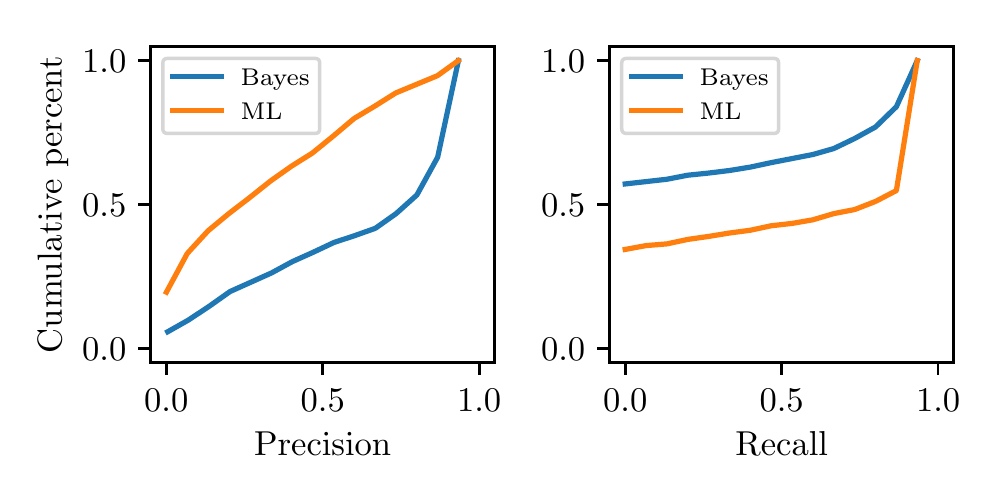}}
}{%
  \caption{Cumulative distribution function for precision and recall of class INFO}%
  \label{fig:cdf-info}
}
\end{floatrow}
\end{figure}

The empirical cumulative distribution functions (CDFs) of class PERSON for precision and recall can be found in \cref{fig:cdf-pers}. Let $F_1$ and $F_2$ be two CDFs, then $F_1$ is \textit{dominated stochastically to 1st order} by $F_2$ \cite{Pflug-et-al-1996},
\begin{equation}
F_1 \prec F_2,\ \text{if}\ F_1(x) \geq F_2(x)\ \forall\ x.
\end{equation}
In the following, we denote the CDFs of the Bayes decision rule regarding precision and recall by $F_{\textit{B}}^p$ and $F_{\textit{B}}^r$, respectively. Analogously, $F_{\textit{ML}}^p, F_{\textit{ML}}^r$ refer to the ML decision rule.

As to be expected, we observe a clear advantage of Bayes in terms of precision since $F_{\textit{ML}}^p \prec F_{\textit{B}}^p$. For any precision value $x$, in particular for low precision values, the frequency that one instance's precision is below $x$ is significantly less with Bayes than with ML. The average difference is about $25\%$. Hence, Bayes predicts PERSON segments with better precision than ML.

In terms of recall, we observe the opposite behavior: $F_{\textit{B}}^r \prec F_{\textit{ML}}^r$, i.e., ML is superior over Bayes in this metric. The average difference is about $23\%$. However, for both decision rules the number of non-detected segments, i.e., $F_{\textit{B}}^r(0)=0.36$ and $F_{\textit{ML}}^r(0)=0.17$, is quite high.

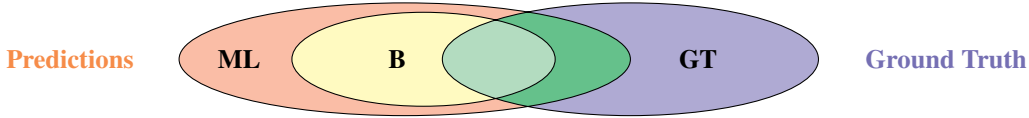
\begin{figure}[]
\centering
\begin{tikzpicture}
    \def\firstellipse{(-2,0) ellipse (3 and 0.75)}
    \def\secondellipse{(1,0) ellipse (2.5 and 0.75)}
	\def\thirdellipse{(-1.75,0) ellipse (1.75 and 0.625)}
	
    \fill[Red!30!white] \firstellipse;
    \fill[Blue!30!white]  \secondellipse;
    \fill[Yellow!30!white]  \thirdellipse;

    \begin{scope}
        \clip \firstellipse;
        \fill[Green!60!white] \secondellipse;
    \end{scope}

    \begin{scope}
        \clip \thirdellipse;
        \fill[Green!30!white] \secondellipse;
    \end{scope}

    \draw \firstellipse \secondellipse \thirdellipse;
    
    \node [align=center] at (-4.2,0) {\textbf{ML}};
    \node [align=center] at (-2.1,0) {\textbf{B}};
    \node [align=center] at (1.9,0) {\textbf{GT}};
 
	\node [align=center] at (5.2,0) {\color{Blue!60!white}\textbf{Ground Truth}};
	\node [align=center] at (-6.45,0) {\color{Orange!90!white}\textbf{Predictions}};
	
\end{tikzpicture}
\caption{Graphical illustration of the relation between Bayes and ML prediction segments for rare classes. The ellipse \textbf{ML} denotes an segment predicted with ML, analogously \textbf{B} denotes the prediction of the same segment with Bayes. Note that $\text{B}\subseteq \text{ML}$. The ellipse \textbf{GT} denotes the corresponding ground truth object. The intersections colored green indicate the true positive pixel predictions, i.e., light green corresponds to the Bayes prediction and dark green to the additional correct pixel predictions obtained with ML. The red color indicates the additional false positive pixel predictions obtained with ML.
}
\label{fig:bay-ml-rel}
\end{figure}

Qualitatively, we observe very similar results for the INFO class (as with PERSON), see \cref{fig:cdf-info}.
In our studies of precision and recall we also observe that the findings from \cref{fig:seg-dif} hold statistically,
i.e., we observe for rare classes that all predicted Bayes segments lie entirely inside of predicted ML segments. A graphical illustration of this is given in \cref{fig:bay-ml-rel}.

\paragraph{False-detection vs. Non-detection.}
The benefit from applying ML instead of Bayes lies mainly with the reduction of non-detected ground truth objects. Therefore, it is reasonable to analyze the quantity of the latter, especially in relation to the amount of predicted false positive segments. Additionally, we analyze the false and non-detection frequencies in relation to the size of the predicted segment and the actual ground truth segment, respectively.

\begin{figure}
\vspace{-.5cm}
\begin{floatrow}
\ffigbox{%
  \centerline{\includegraphics[trim={0 0 0 0.5cm},width=.49\textwidth]{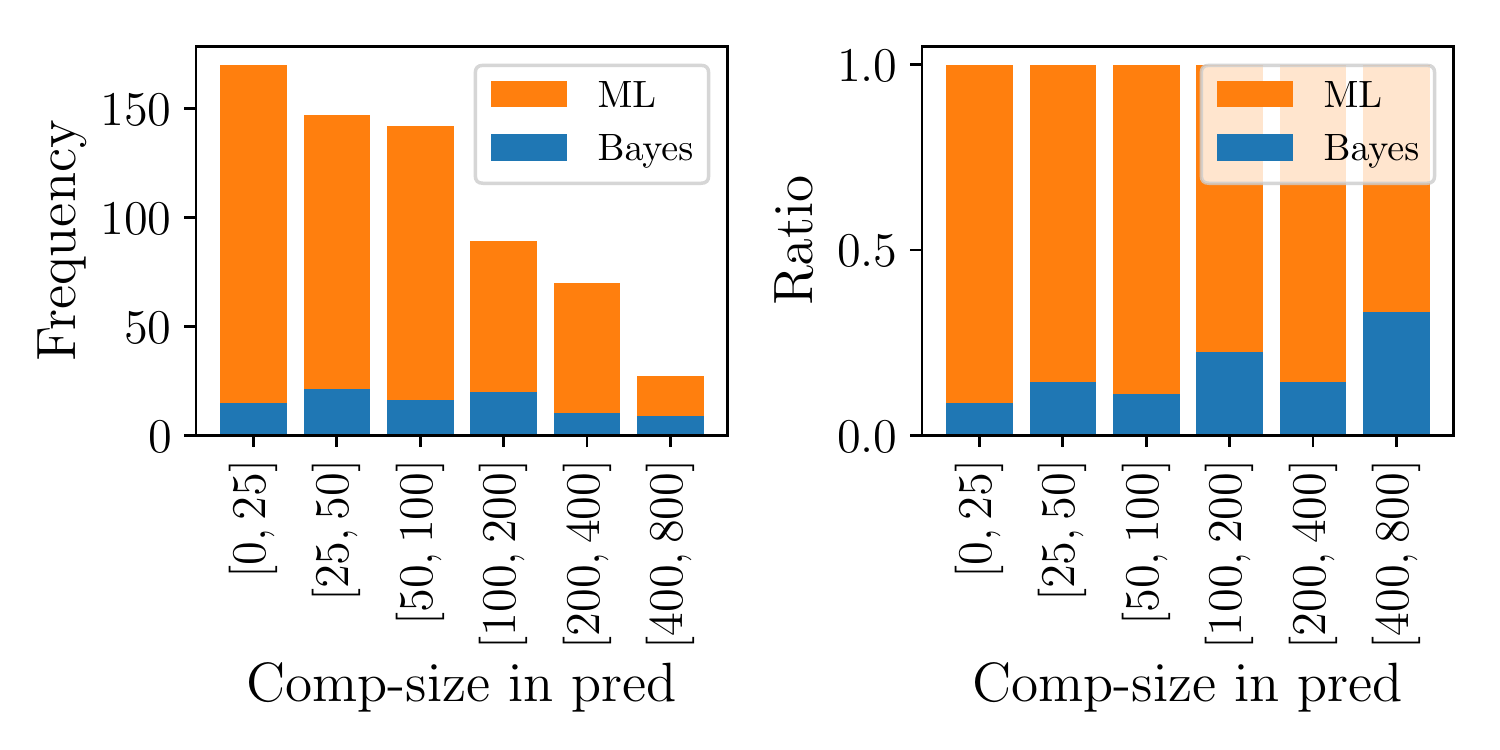}}
}{%
  \caption{False-detection of class PERSON conditioned on segment size in prediction}%
  \label{fig:fd-pers}
}
\ffigbox{%
  \centerline{\includegraphics[trim={0 0 0 0.5cm},width=.49\textwidth]{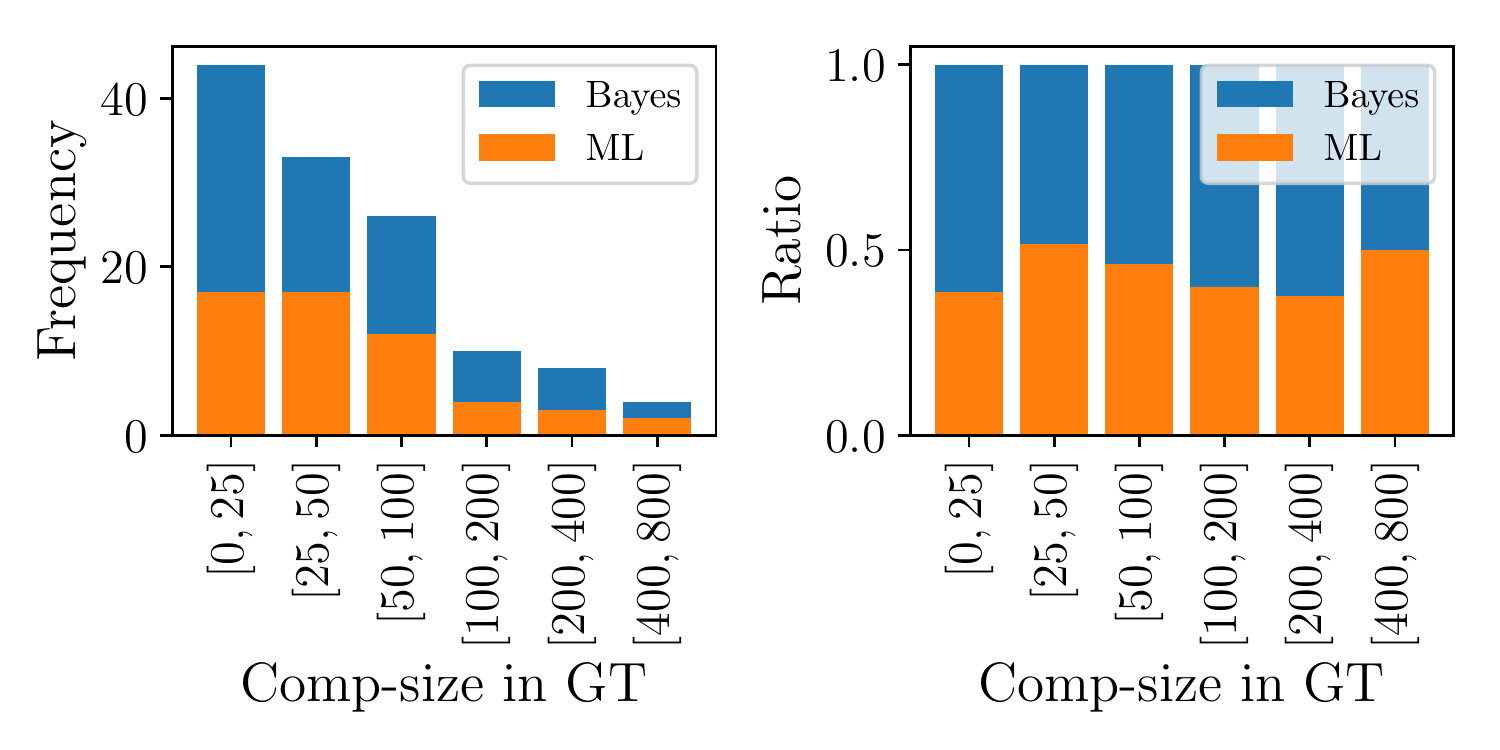}}
}{%
  \caption{Non-detection of class PERSON conditioned on object size in ground truth}%
  \label{fig:nd-pers}
}
\end{floatrow}
\end{figure}

\Cref{fig:fd-pers} depicts the number of false positive PERSON segments depending on the size of the instance in the segmentation for ML and Bayes. In the left panel, there is a noticeable decrease in the frequency of false positive ML segments if the segment size increases, i.e., larger predicted segments are less likely to be entirely incorrect. For Bayes segments, the same tendency holds, even though not as strictly as for ML. Moreover, we see for every segment size bin that the amount of false positive ML segments considerably exceeds the amount of false positive Bayes segments. The right-hand panel of \cref{fig:fd-pers} shows the amount of Bayes false positives relative to the amount of ML false positives for different component sizes. For increasing component size there is also an increase in the relative amount of Bayes false positives.

\begin{figure}
\vspace{-.25cm}
\begin{floatrow}
\ffigbox{%
  \centerline{\includegraphics[trim={0 0.5cm 0 0},width=.49\textwidth]{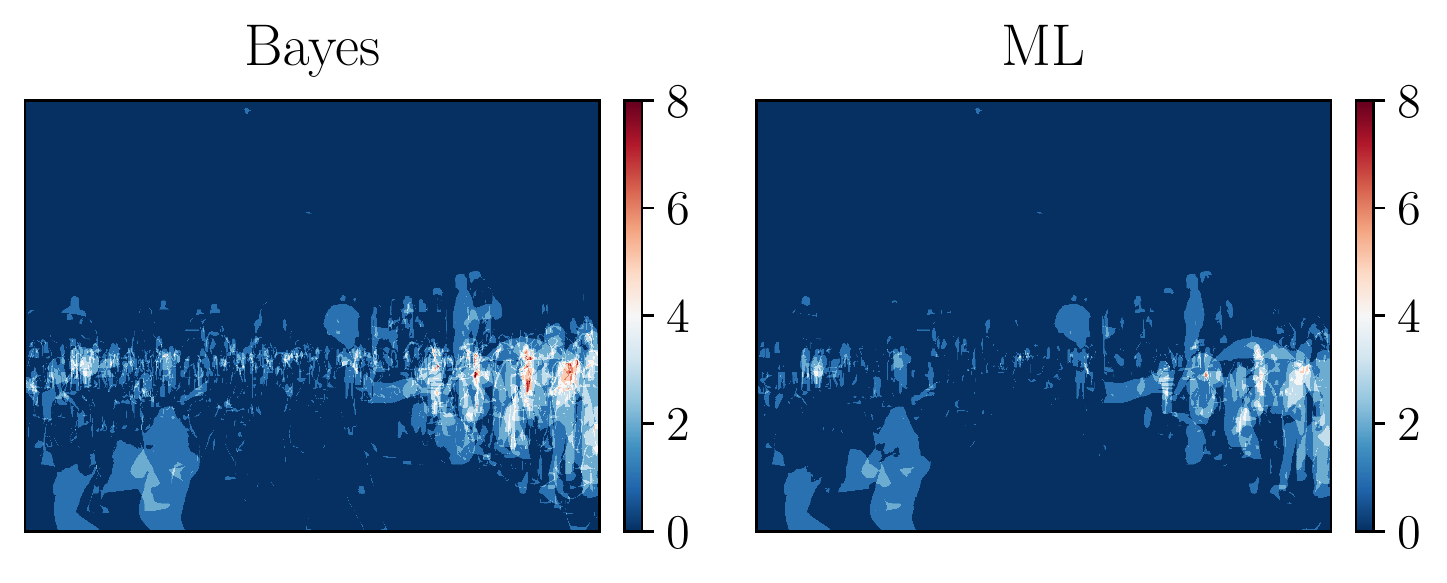}}
}{%
  \caption{Heat map of non-detected pixels of class PERSON}%
  \label{fig:nd-px-pers}
}
\ffigbox{%
  \centerline{\includegraphics[trim={0 0.5cm 0 0},width=.49\textwidth]{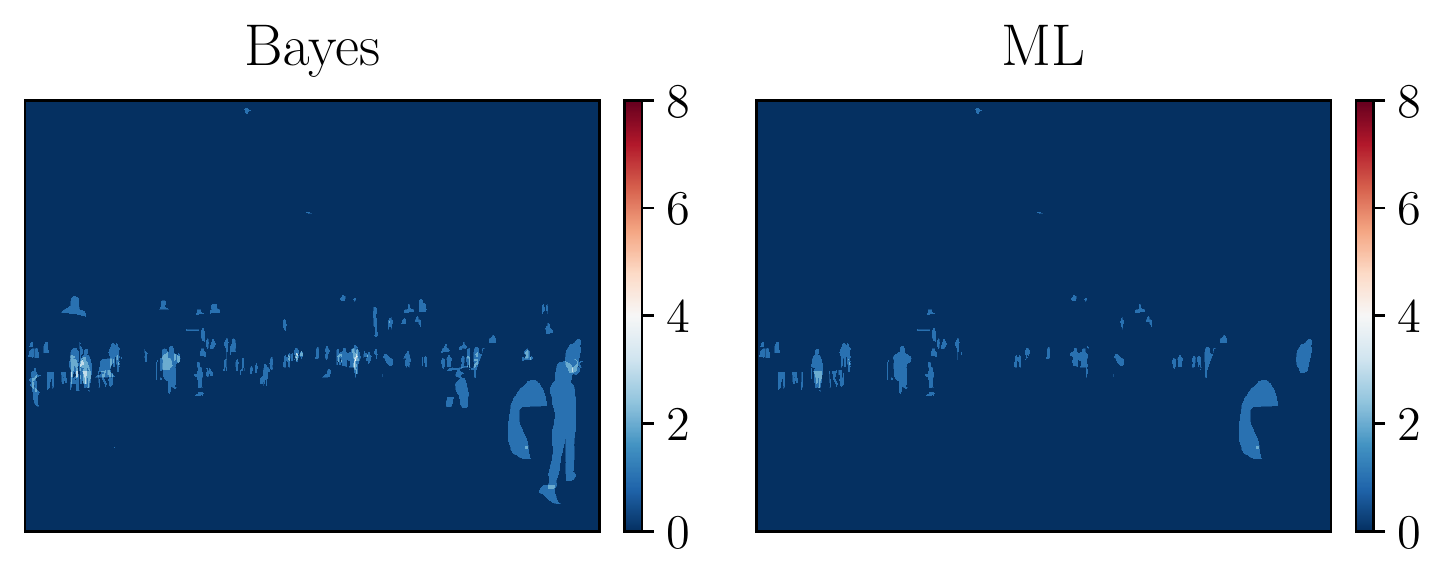}}
}{%
\caption{Heat map of non-detected objects of class PERSON}%
  \label{fig:nd-obj-pers}
}
\end{floatrow}
\end{figure}

Analogous to \cref{fig:fd-pers}, \cref{fig:nd-pers} shows the number of entirely non-detected objects depending on their size. We observe a similar behavior for Bayes and ML like for the false-detections, also with respect to the object size. For the non-detection of the class PERSON we find a clear advantage in favor of ML, independent of the object size. Bayes overlooks roughly twice as many objects as ML does. This result indicates an uncertainty of the network in finding the rare class PERSON which can be alleviated by using ML. \Cref{fig:nd-px-pers} and \cref{fig:nd-obj-pers} visualize the non-detection at pixel and at object level, respectively.

For the class INFO we observe an analogous behavior. We refer to the appendix for the analysis and corresponding figures.

\section{Conclusion and Outlook} \label{conclusion}

In this work, we conducted an in-depth comparison of the ML and Bayes decision rules for a semantic segmentation network trained on an unbalanced dataset showing street scenes. In our tests we observe that ML is able to detect the rare classes PERSON and INFO more frequently than Bayes. Indeed, the pixels that Bayes and ML classify differently indicate that a less frequent class might be overlooked. We have seen that ML detects significantly more instances of rare classes in comparison to Bayes, but to the the detriment of producing even substantially more false-detections which makes ML not reliable for always predicting rare classes correctly. Apart from this, it is important to emphasize that ML only post-processes the softmax output of a neural network. This can be done simultaneously while applying the usual Bayes rule. In the end, we obtain two prediction masks and the additional ML mask is produced computationally nearly for free. What remains is to develop methods to draw plausible conclusions in order to combine both segmentations. Furthermore, the ML prediction can serve as uncertainty mask revealing labeling mistakes of training data or indicating new unlabeled images of high prediction uncertainty which then can be annotated and included in the training process in the manner of active learning. We make the source code of our analysis tool publicly available on GitHub: \texttt{\href{https://github.com/robin-chan/decision-rules}{https://github.com/robin-chan/decision-rules}}.

\paragraph{Acknowledgements.} This work is funded in part by Volkswagen Group Research.

\printbibliography

\section*{Appendix}

\paragraph{False-detection vs. Non-detection for INFO.}

\begin{figure}
\begin{floatrow}
\ffigbox{%
  \centerline{\includegraphics[trim={0 0 0 0.5cm},width=.49\textwidth]{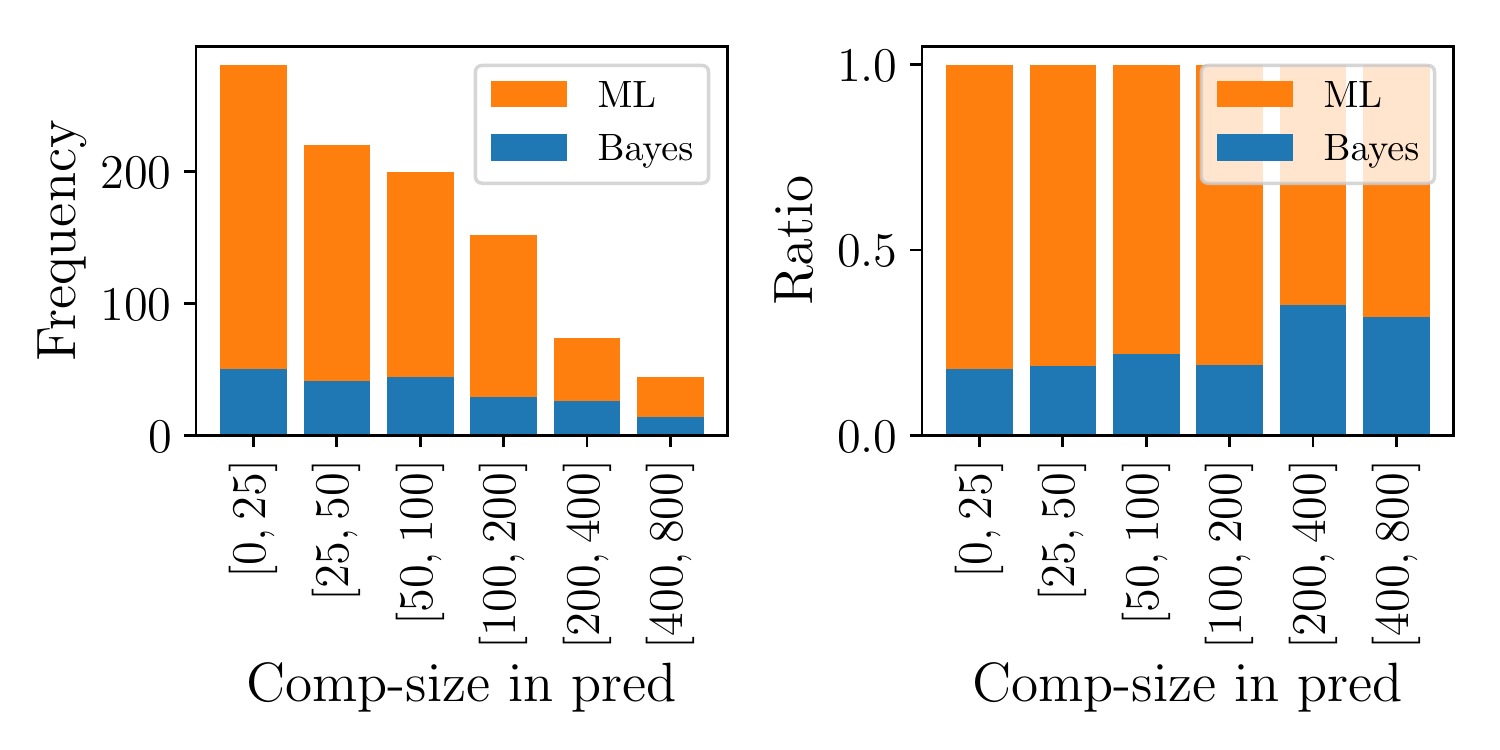}}
}{%
  \caption{False-detection of class INFO conditioned on instance size in prediction}%
  \label{fig:fd-info}
}
\ffigbox{%
  \centerline{\includegraphics[trim={0 0 0 0.5cm},width=.49\textwidth]{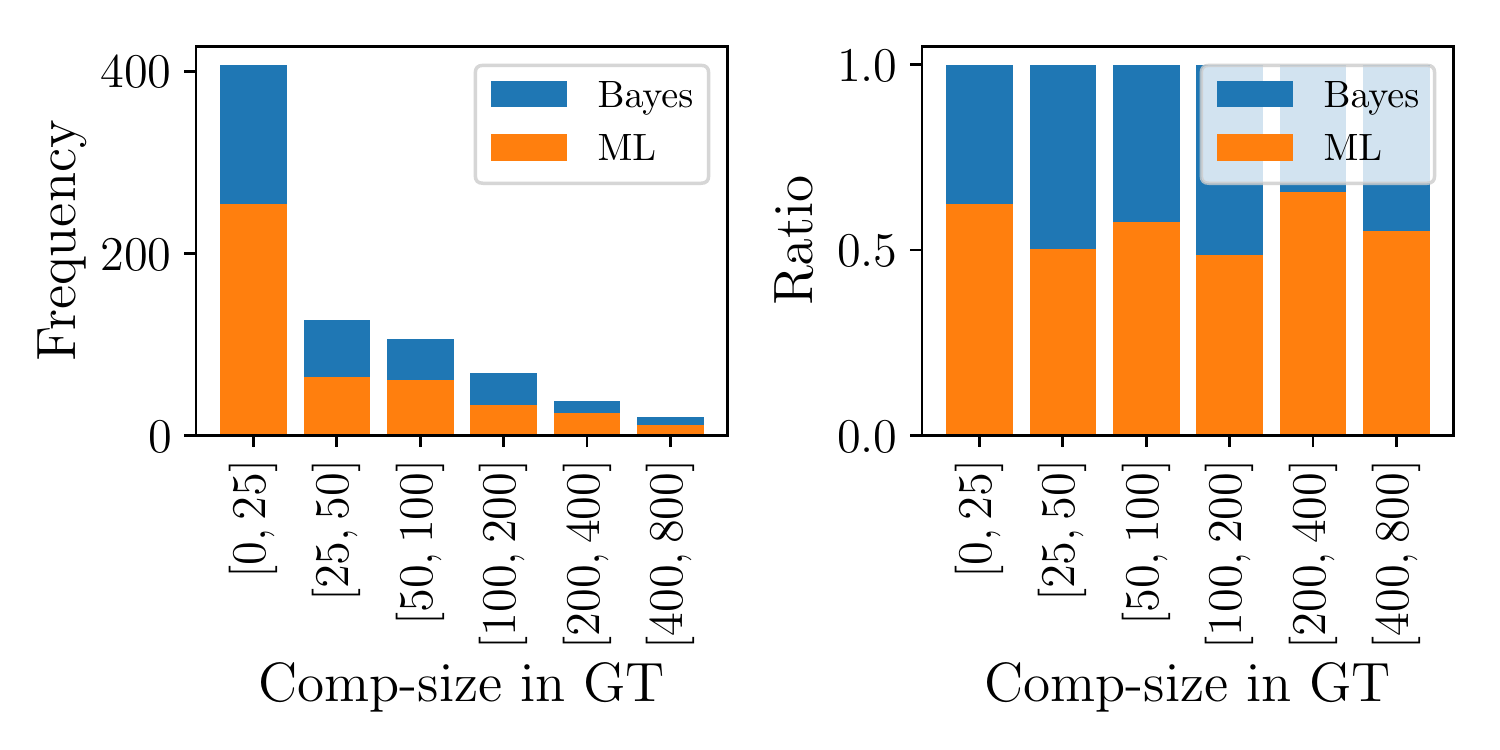}}
}{%
  \caption{Non-detection of class INFO conditioned on object size in ground truth}%
  \label{fig:nd-info}
}
\end{floatrow}
\end{figure}

Investigating the amount of false-detections as shown in the left panel of \cref{fig:fd-info}, we notice a staircase-like decrease of false positive ML and Bayes segments for increasing segment size. Moreover, in general there are also more false-detections with ML than with Bayes which can be studied comprehensively in terms of the ratio of false-detections, see right panel of \cref{fig:fd-info}. On average the amount of Bayes false positives is about $23\%$ of the amount of ML false positives. A slight increase of this percentage can be observed for larger component sizes.

\begin{figure}
\begin{floatrow}
\ffigbox{%
  \centerline{\includegraphics[trim={0 0.5cm 0 0.25cm},width=.49\textwidth]{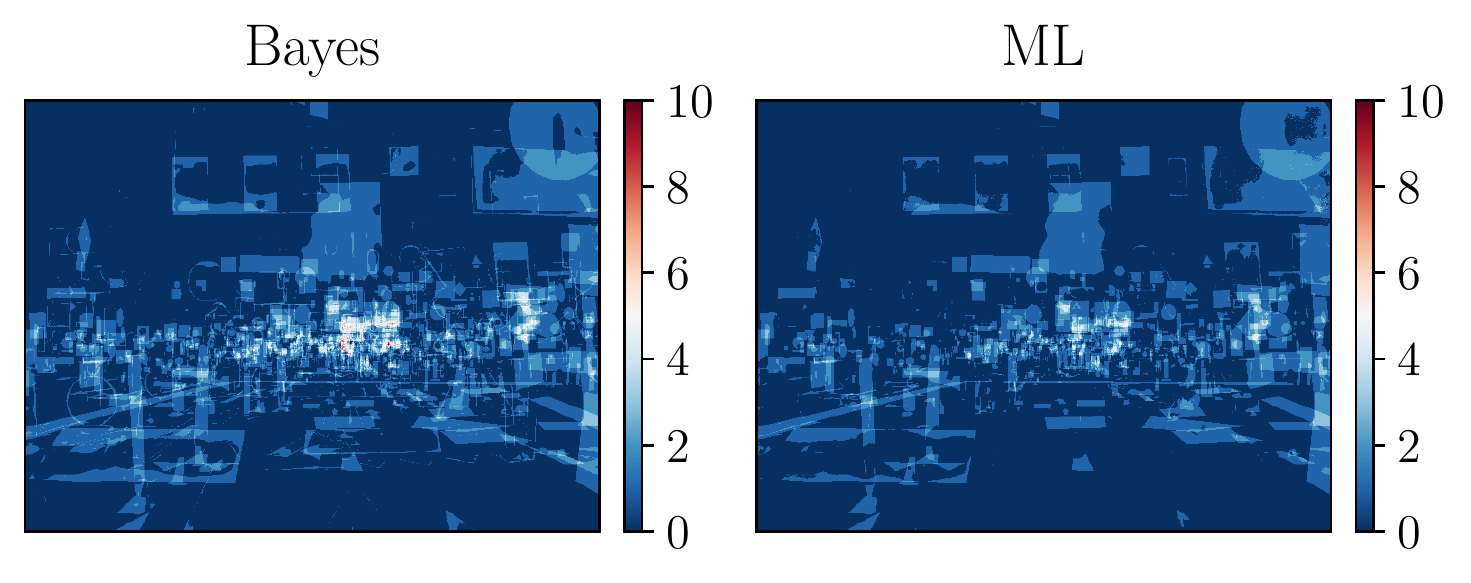}}
}{%
  \caption{Heat map of non-detected pixels of class INFO}%
  \label{fig:nd-px-info}
}
\ffigbox{%
  \centerline{\includegraphics[trim={0 0.5cm 0 0.25cm},width=.49\textwidth]{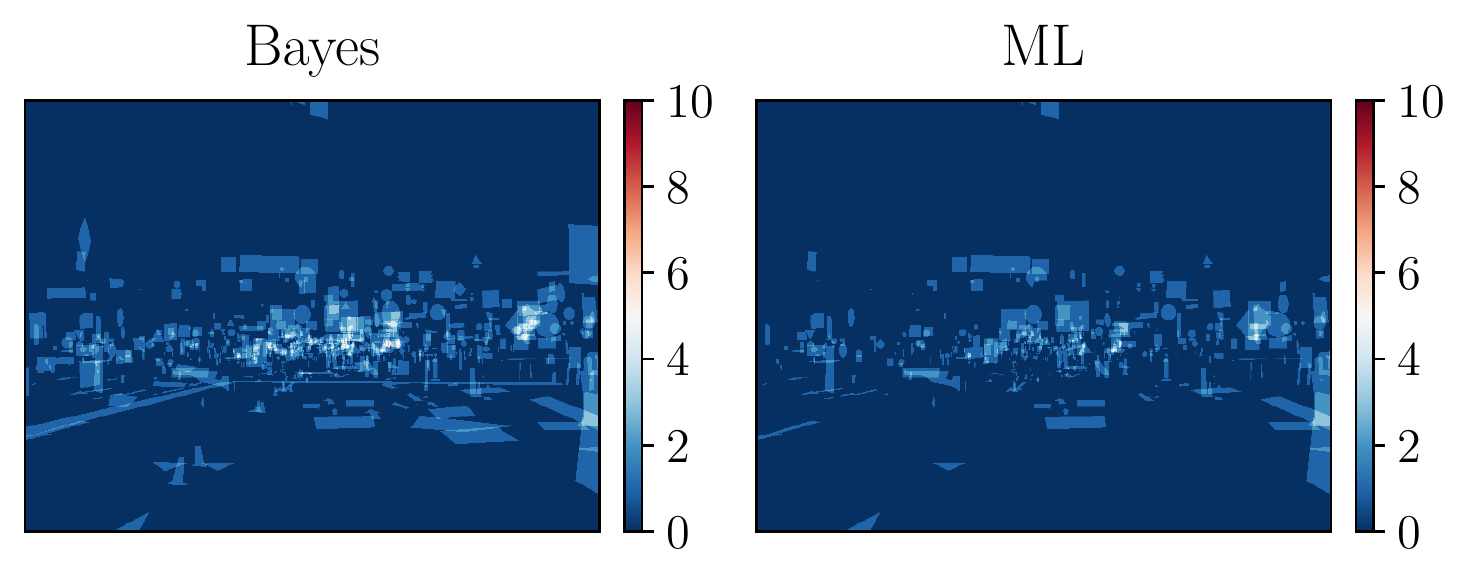}}
}{%
  \caption{Heat map of non-detected objects of class INFO}%
  \label{fig:nd-obj-info}
}
\end{floatrow}
\end{figure}

The left panel of \cref{fig:nd-info} shows that less info signs are entirely overlooked with ML than with Bayes. For both rules the frequency decreases for larger objects. Nevertheless, most objects of the considered class are rather small which noticeably affects the frequency of non-detections. From the ratio in the right panel of \cref{fig:nd-info}, we conclude that Bayes fails to detect approximately every second info sign, while ML fails to detect the same every third time. Compared to the performance for class PERSON, ML does not outperform Bayes on detecting info signs as greatly as on detecting persons, but still provides a remarkable performance gain. A visualization of the just mentioned benefit of ML is given in \cref{fig:nd-px-info} and \cref{fig:nd-obj-info}.

\paragraph{Local Priors vs. Global Prior.}

\begin{figure}
\vspace{-0.5cm}
\begin{floatrow}
\capbtabbox[.37\textwidth]{%
\scalebox{.93}{
\begin{tabular}{l|c}
\toprule
Class       & Global Prior  \\ \midrule
PERSON      & 0.0022        \\ 
INFO SIGN   & 0.0067        \\ \bottomrule
\end{tabular}

}
\vspace{.2cm}
}{\caption{Global prior of class PERSON and INFO which is the average value over all pixel-wise priors of the respective class, i.e., the proportion of all pixels in the training set belonging to that class (see also \cref{fig:ds20k-imbalance}).}%
\label{table: glob-prior}
}

\ffigbox[.61\textwidth]{%
  \centerline{
  \begin{minipage}{.2\textwidth}
  \centering
  \includegraphics[width=.9\textwidth]{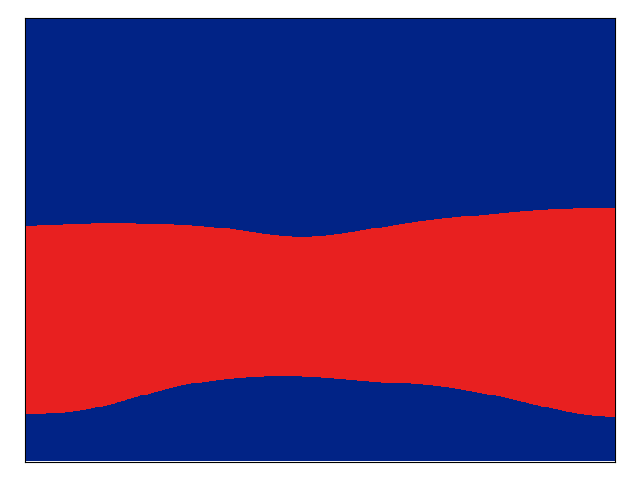}
\end{minipage}%
\begin{minipage}{.2\textwidth}
  \centering
  \includegraphics[width=.9\textwidth]{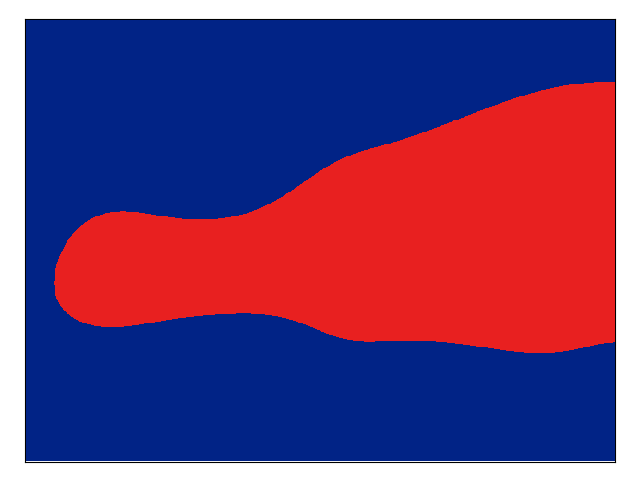}
\end{minipage}
\begin{minipage}{.2\textwidth}
  \centering
  \includegraphics[width=.9\textwidth]{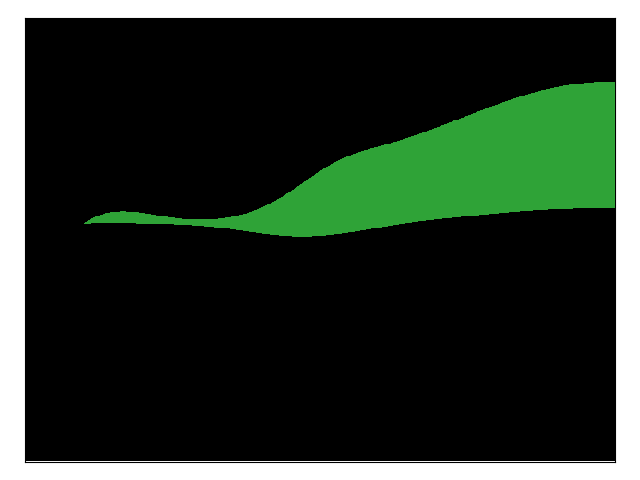}
\end{minipage}
}%
}{%
  \caption{Visualization of the two sets $B_k$ (red) and $B^\prime_k$ (blue) for the classes PERSON (left panel) and INFO (center panel). In addition, the green color (right panel) shows $B_{\textit{PERSON}}^\prime \cap B_{\textit{INFO}}$, i.e., the region of pixel positions in which the local priors are lower than the global prior for class the PERSON and the other way round for INFO.}%
  \label{fig:heat-dif}
}
\end{floatrow}
\end{figure}

In \cref{sec:decision-rules-nn}, we introduced a method that uses pixel-wise ``local'' priors in order to handle the class imbalance depending on the location in the image. However, one might argue that a fully convolutional segmentation network is translation invariant (up to some stride caused by pooling) and the choice of priors should take this into account. We justify our preference over non-localized ``global'' priors with the network's large receptive field.
To this end, we illustrate the impact regarding the non-detection of a PERSON instance when using local priors in comparison to a global prior, see \cref{table: glob-prior}, in particular when the local priors are lower than the global prior. 

For class $k = 1, \ldots, N$, let $p^g(k)$ be the global prior and let $p_{ij}^l(k)$ be the local prior at pixel position $(i,j) \in \{1,\ldots,m\} \times \{1,\ldots,n\}$. Then, we denote by
\begin{align}
    B_k := \{ (i,j) : p^g(k) \leq p_{ij}^l(k) \} \\
    B_k^\prime := \{ (i,j): p^g(k) > p_{ij}^l(k) \}
\end{align}
the set of pixel positions where the global prior is lower than or equal to the local prior and the set of pixel positions where the global prior is higher than the local prior, respectively.

For our test, we place one PERSON instance at $(i,j) \in B_{\textit{PERSON}}^\prime$, see \cref{fig:heat-dif} left image. Since, in this region of the image the network rather expects to see an info sign and the local priors of class PERSON and INFO are of a similar magnitude, we cropped the image such that the PERSON instance is located at $(i,j) \in B_{\textit{PERSON}}^\prime \cap B_{\textit{INFO}}$ in order to provoke a misclassification by using global priors, see \cref{fig:heat-dif} center and right image.

\begin{figure}
\input{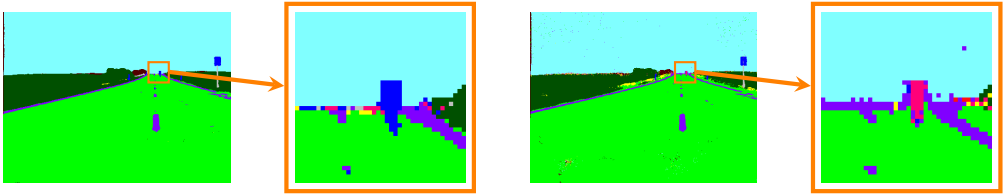}
\caption{Example of a non-detected person by using a global prior (two left hand panels) which is detected by using local priors (two right hand panels).}
\label{fig:global-vs-local}
\end{figure}

\Cref{fig:global-vs-local} shows the segmentations produced by using the different priors. We observe that the person is entirely overlooked and predicted to be an info sign by using the global prior while the person is nearly fully detected by using the local priors. Although, this situation is artificially created, it illustrates the importance of using priors and, in particular, the positive effect of localized priors for images outside of the network's learned experience.
\end{document}